\pdfoutput=1

\documentclass[11pt]{article}

\usepackage{acl}

\usepackage{times}
\usepackage{latexsym}
\usepackage{graphicx}
\usepackage{xcolor}

\usepackage[T1]{fontenc}

\usepackage[utf8]{inputenc}

\usepackage{microtype}

\title{Unsupervised Sentence Simplification via Dependency Parsing}

\author{Vy Vo \quad Weiqing Wang \quad Wray Buntine \\
  Faculty of Information Technology, Monash University, Australia \\
  \texttt{tvoo0019@student.monash.edu} \\ 
  \texttt{\{teresa.wang, wray.buntine\}@monash.edu}
  }

\begin{document}
\maketitle
\begin{abstract}
Text simplification is the task of rewriting a text so that it is readable and easily understood. In this paper, we propose a simple yet novel unsupervised sentence simplification system that harnesses parsing structures together with sentence embeddings  to produce linguistically effective simplifications. This means our model is capable of introducing substantial modifications to simplify a sentence while maintaining its original semantics and adequate fluency. We establish the unsupervised state-of-the-art at 39.13 SARI on TurkCorpus set and perform competitively against supervised baselines on various quality metrics. Furthermore, we demonstrate our framework's extensibility to other languages via a proof-of-concept on Vietnamese data. Code for reproduction is published at \url{https://github.com/isVy08/USDP}.
\end{abstract}

\section{Introduction}
Text simplification (TS) contributes to promoting social inclusion by making information more accessible to people with reading comprehension problems such as second-language learners, non-experts or those suffering from cognitive impairment \cite{siddharthan2014survey,stajner-2021-automatic}. Simplification can take many forms. It may involve \textbf{splitting} and/or \textbf{reordering} parts of the text to produce simpler syntactic structures. \textbf{Lexical transformation} can also be performed through the substitution of easier and more familiar vocabularies. While \textbf{deleting} redundant details produces an output shorter in length, the output can also become longer with extra information added to provide explanations for difficult concepts, i.e., \textbf{elaboration}. Regardless of the operations, the simple variants must preserve the key meaning of the original texts. 

\begin{table}[h!]
\centering
\begin{tabular}{p{3.5cm}|p{3.5cm}}
\hline
\textbf{Original sentence} & \textbf{Simple variant} \\ [0.5ex]
\hline
In Ethiopia, HIV disclosure is low & In Ethiopia , HIV is low \\
\hline
Mustafa Shahbaz , 26 , was shopping for books about science & Mustafa Shahbaz, 26 years old , was a group of books about science \\
\hline
Healthy diet linked to lower risk of chronic lung disease & Healthy diet linked to lung disease \\
\hline
\end{tabular}
\caption{Examples of logic errors produced by \texttt{ACCESS} \cite{martin-etal-2020-controllable} and \texttt{DMLMTL} \cite{guo2018dynamic}, taken from \cite{garbacea-etal-2021-explainable}.}
\label{tab1}
\end{table}

Previous studies criticize existing systems for being opaque, suboptimal and semantically compromising \cite{garbacea-etal-2021-explainable,NAACL-2021-Maddela,stajner-2021-automatic}. Table \ref{tab1} illustrates how the meaning is easily sacrificed with respect to minor changes to the sentence structure. 
Generally, TS works must be mindful of the trade-off of \textbf{Simplicity} vs. \textbf{Fluency} and \textbf{Adequacy} (semantics similarity). Although some works explicitly incorporate all three properties in the training objectives \cite{laban2021keep_it_simple,zhang-lapata-2017-sentence,kumar-etal-2020-iterative}, their evaluations do not clearly explain whether or not simplicity is induced as the cost of fluency and adequacy. By exploiting deep dependency parsing, we contribute a novel unsupervised strategy called \textbf{Family Sampling} that strictly enforces grammatically fluent simplification while ensuring the most important ideas are retained. We shed light on how our model achieves the balance of these three properties through both automatic metrics and human judgement, which at the same time substantiates our superiority over unsupervised counterparts.

Whereas most models are restricted to the language of the data they are trained on, we demonstrate that our framework readily extends to other languages by adapting it to simplify Vietnamese texts. We also address interpretability by providing linguistically-motivated empirical evidence confirming the intuition behind our framework.  

\section{Related Work}
\textbf{Supervised TS}. Earlier works inherit techniques from statistical machine translation \cite{brown-etal-1990-statistical} to translate a text of the \textit{complex} language to the \textit{simple} language. The translation model is learned through aligned words or phrases in normal-simplified text pairs, referred to as phrase-based simplification \cite{coster-kauchak-2011-simple,koehn-etal-2007-moses,narayan-gardent-2014-hybrid,wubben-etal-2012-sentence}. Alternatively, in syntax-based simplification \cite{zhu-etal-2010-monolingual}, the alignment units are syntactic components. The first neural sequence-to-sequence text simplification system is proposed in \cite{nisioi-etal-2017-exploring}. Utilizing the same architecture, other works \cite{guo2018dynamic,zhang-lapata-2017-sentence} further employ reinforcement learning with a reward function that is a weighted sum of three component rewards: simplicity, relevance and fluency. Audience Centric Sentence Simplification \texttt{ACCESS} is a recent supervised state-of-the-art approach 
\cite{martin-etal-2020-controllable} that conditions the simplified outputs on different attributes of text complexity. These models rely on parallel corpora to implicitly learn hybrid transformation patterns. Despite impressive results, the scarcity of high-quality and large-scale datasets heavily impedes progress in supervised TS \cite{alva-manchego-etal-2020-asset}. The attention has thus been shifted towards semi-supervised and unsupervised approaches.

\textbf{Semi-Supervised TS}. Instead of using aligned data, \citet{zhao2020semi} introduce a noising mechanism to generate parallel examples from any English dataset, then train denoising autoencoders to reconstruct the original sentences. Under a framework called \textbf{back-translation}, \citet{martin2021muss} use multilingual translation systems to produce various simpler paraphrases from monolingual corpora (e.g., English to French, then French to English), thus eliminating the need for labeled data. In the same spirit, \citet{mallinson-etal-2020-zero} recently adopt a cross-lingual strategy to simplify low-resource languages (e.g., German) from high-resources languages (e.g., English). Instead of having translation done in the data pre-processing step, they develop a system that performs simplification and translation at the same time. 

\textbf{Unsupervised TS}. The first unsupervised neural model for text simplification is proposed by \citet{surya-etal-2019-unsupervised}. The model is trained to minimize adversarial losses on two separate sets of complex and simple sentences extracted from a parallel Wikipedia corpus. \texttt{DisSim} \cite{niklaus-etal-2019-transforming} is another effort focusing on splitting and deletion by applying 35 hand-crafted grammar rules over a constituency parse tree. Recent works tend to favor edit and decoding-based approaches. This line of work is advantageous since not only can the system generate hybrid outputs without relying on aligned datasets, but it can also allow for quality control explicitly via a scoring function balancing simplicity, fluency and semantics preservation. The algorithm of \citet{kumar-etal-2020-iterative} iteratively edits a given complex sentence to make it simpler using four operations: removal, extraction, reordering and substitution, while \cite{kariuk2020cut} implement beam search with simplicity-aware penalties for sentence simplification. In the same setting as \cite{zhang-lapata-2017-sentence}, \texttt{KiS} \cite{laban2021keep_it_simple} revisits reinforcement learning and tackles simplification for paragraphs without supervision. However, the method involves end-to-end training on multiple Transformer-based models, which is computationally expensive and challenging to extend to new settings. 

\textbf{Contributions}. In contrast to the above, we propose a light-weight solution for sentence TS making full use of pre-training, i.e., neither fine-tuning nor end-to-end training is required, thereby making it highly reproducible and robust to out-of-distribution examples. 
Specifically, we improve on the existing decoding procedure through a linguistics-based unsupervised framework. We perform structural and lexical simplification sequentially, rather than simultaneously like previous works since it would support interpretation and allow for controllability. This sequential approach is also adopted in \cite{NAACL-2021-Maddela}, which leverages \texttt{DisSim} together with a self-designed paraphrasing system. We first develop a \textbf{stand-alone decoding framework} for structural simplification, then adopt \textbf{back-translation} for lexical simplification and paraphrasing. After studying prior works, we find that \textbf{splitting} and \textbf{elaboration} are difficult to implement without labeled data or heavily injected grammar rules, whereas our goal is to maximize the capacity of TS system in the absence of external knowledge. An interesting discovery is that back-translation in phase $2$ is a convenient technique, in that it can also perform \textbf{reordering} (as part of the rewriting process), if the operation is necessary to produce a familiar structure. Thus, in the first phase, we choose to focus on \textbf{deletion} - the operation that easily leads to poor adequacy if not carefully done. 
Although we only tackle sentence simplification, in \textbf{Section 7}, we provide directions on how our framework can be flexibly adapted for various purposes, including paragraph-level simplification.
\section{Method}
We propose a two-phase pipeline that  tackles syntactic and lexical simplification one by one. The first phase consists of an independent left-to-right decoder operating in a much more efficient search space induced by dependency relations among words in a sentence. Though this phase is mainly about deletion, during the process, the system can also perform \textbf{chunking}, i.e., breaking a sentence into meaningful phrases. In the second phase, we back-translate the generated English outputs from phase $1$ to generate effective paraphrases and lexical simplifications. 
\subsection{Structural Simplification} 
\subsubsection{Search Objective} 
Given an input sentence $c := (c_1, c_2, ..., c_n)$, we aim to generate a shorter sentence $s := (s_1, s_2, ..., s_m)$ expressing the same meaning as $c$. Whereas the previous works perform deletions by imposing length constraints, we go beyond length reduction and strictly define which parts of a sentence to keep and which to remove. The goal is to eliminate redundant details – those if removed do not significantly alter the meaning of the entire sentence. We quantify the importance of a word by measuring local changes in semantic similarity scores when omitting it from the original sentence. 
This motivates our search objective function as follows
\begin{equation} \label{eq1}
f(s) = f_{sim}(c, s) +  \alpha f_{flu}(s) + f_{depth}(s)
\end{equation}
\noindent
where $\alpha$ is the relative weight on Fluency score. The reason why the weights of score functions $f_{sim}$ and $f_{depth}$ are the same is explained under \textbf{Section 4.2}. The decoding objective is a linear combination of individual scoring functions with each score normalized within the range $[0,1]$. Details of each function are described below.

\textbf{Semantic Similarity $f_{sim}(c,s)$.} Using cosine distance as a proxy for semantic relevance has been widely adopted across TS works. We calculate cosine similarity between sentence embeddings of $c$ and generated hypothesis $s_{1:t}$ at each time step $t$. We utilize the pre-trained sentence-BERT model (SBERT) \cite{reimers-gurevych-2020-making}, which is best known for its superiority in producing semantically meaningful sentence embeddings, whereas other unsupervised works use weighted average of word embeddings \cite{kumar-etal-2020-iterative,schumann-etal-2020-discrete,zhao2020semi} or LSTM-encoded hidden representations \cite{zhang-lapata-2017-sentence}. 
$$f_{sim}(c, s_{1:t}) = cos(e_c, e_{s_{1:t}})$$  
The intuition is, if a candidate word $i$ is more important than another word $j$, add $i$ to the sentence will increase the similarity score more than when adding $j$. It is observed that SBERT sentence embeddings capture this behavior, and exactly how it works is explained in \textbf{Section 6}. 

\textbf{Fluency $f_{flu}(s)$.} Our fluency scorer quantifies the grammatical accuracy of a sequence based on a constituent-based 4-gram language model, given as
$$ f_{flu}(s_{1:t}) = \frac{1}{|s_{1:t}|} \sum_{u=1}^t \log p(pos_u\,|\,pos_{1:u-1})$$ where $pos_t$ indicates the part-of-speech of token $s_t$ The language model is pre-trained on a massive unlabeled corpus. Because English constituents are bounded, constituent-based language model is a reusable light-weight solution compared to regular vocabulary-based language models. 

\textbf{Tree Depth Constraint $f_{depth}(s)$.} Dependency tree depth is a popular measure of syntactic complexity in various literature in linguistics \cite{genzel-charniak-2003-variation,sampson1997depth,xu-reitter-2016-convergence}. A deeper tree contains more dependencies indicating complex structures e.g., usage of subordinate clauses. \texttt{ACCESS} \cite{martin-etal-2020-controllable} has recently provided empirical evidence showing that controlling maximum depth of dependency tree yields the most effective syntactical simplifications. Thus, $f_{depth}$ further scores candidate sentences by the \textbf{inverse maximum tree depth} reached at the generated token. This constraint prevents the decoder from going too deep, thereby producing a structurally simpler output.    

\subsubsection{Search Space} 
Deletion is a form of extractive summarization by nature, motivating us to adopt the word-extraction method proposed by \cite{schumann-etal-2020-discrete}. They suggest candidates be selected from tokens in the input sentence, instead of the corpus vocabulary. Specifically at each step, a new candidate is sampled from words in the input sentence that are not in the current summary. We argue that this is not necessary and propose a more efficient approach. Each token exists in a directed relation with its parent (or head), which determines the grammatical function of that token. The head-dependent relation is also an approximation for the semantic relationship between them \cite{jurafsky2014speech}. At each step, we thus only consider tokens that are the children of the previously generated token, resulting in a smaller search space. We additionally arrange the words in the same order as the input before scoring hypotheses since word order plays a critical role both grammatically and semantically. This approach allows us to achieve optimal solution while using a small beam size. We refer to this novel strategy as \textbf{Family Sampling}.

\subsubsection{Search Algorithm} 
\begin{figure*}[h!]
\centering
\includegraphics[width=1.0\textwidth]{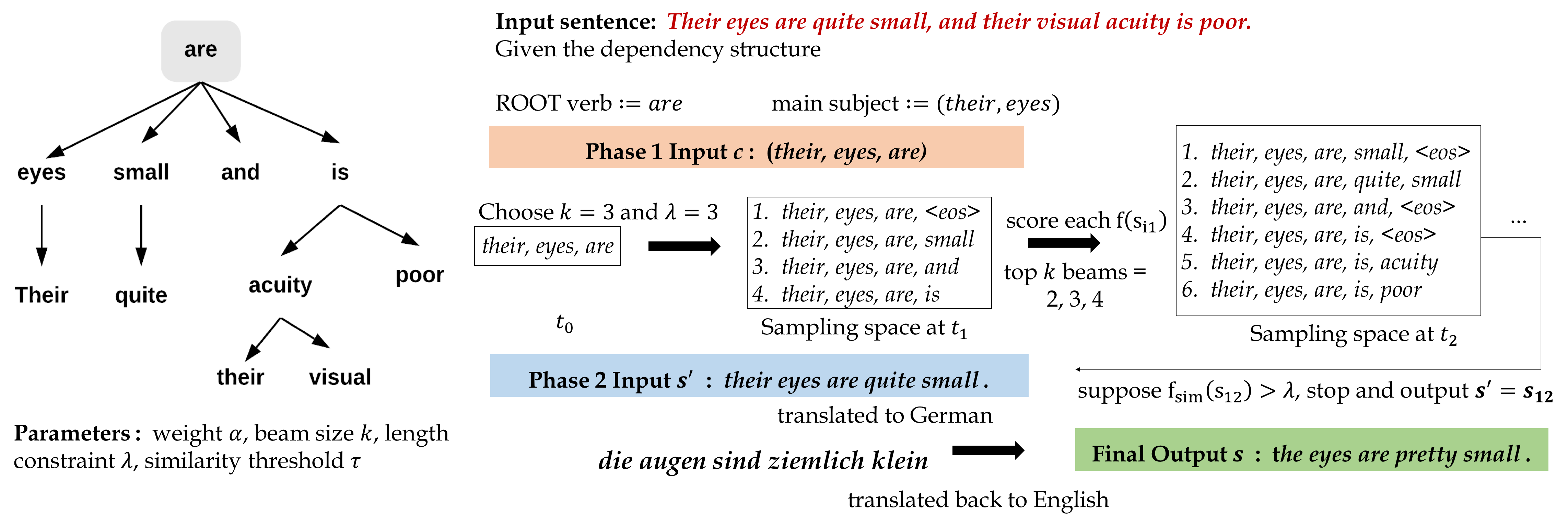}
\caption{At each search step, a candidate token is sampled from the direct children of previously generated token. Score each candidate according to the objective function and select $k$ hypotheses with highest scores for the next generation step. The decoder terminates once there is a sequence reaching length $\lambda$ with at least similarity $\tau$.} \label{fig1}
\end{figure*}

Figure \ref{fig1} provides a running example for our algorithm. We integrate our novel family sampling strategy with a regular beam search algorithm that keeps top $k$ hypotheses with highest scores. To begin with, we condition the sequence on the main subject of the sentence, i.e. the subject of the \texttt{ROOT} verb. This does not affect the rest of the sentence, but contributes to simplification by directly introducing the main verb and subject. For each search step, a candidate token $s_t$ is sampled from the family of child nodes of token $s_{t-1}$, excluding those having been previously generated. We score each candidate according to (\ref{eq1}) and select $k$ hypotheses with the highest scores for the next generation step. Our search terminates when the output sentence reaches a predefined length $\lambda$ and satisfies the minimum similarity threshold $\tau$. We wish to find the shortest most similar sub-sequence, and the intuition is to preserve as much semantics as possible by keeping tokens that add the most semantic value to the sequence. As mentioned in \cite{schumann-etal-2020-discrete}, given input length $n$, target output length $m$ and corpus vocabulary size $V$, auto-regressive or edit-based generation has search space of $V^m$, while ours is restricted to $C^{m}_{n}$. Regarding time complexity, our algorithm is bounded by $O(d \times k \times max\_ch(s_{t-1}))$ with parsing tree depth $d$, beam size $k$ and $max\_ch(s_{t-1}))$ being the maximum number of children of a token $s_{t-1}$ where $ch(s_{t-1}) \leq n$ and 

\textbf{Chunking.} A branch in the tree corresponds to a meaningful chunk in the sentence. A \texttt{<SEP>} token is added to induce a chunk whenever a sampling set is empty. Humans naturally perform simplification in this manner by chunking a complex sentence into understandably simpler structures. Our results show that most of these chunks tend to be prepositional and adjective phrases. After a \texttt{<SEP>}, we reverse the tree and restart family sampling with the token nearest to \texttt{ROOT}. 

\subsection{Back Translation} 
Phase $2$ consists of two reliable off-the-shelf machine translation systems English-$X$ and $X$-English where $X$ can be any other language. We simply translate the structurally simpler sentences in English to $X$, then have the outputs back-translated to English. This technique is applied in style transfer tasks \cite{prabhumoye-etal-2018-style,zhang2018style} to disentangle the content and stylistic characteristics of the text. Thus, we rely on back-translation to strip the complex style off a text while keep meaning unchanged. Not only does it enhance the quality of our simplifications via paraphrasing and lexical substitution, but we also find it particularly useful to correct subpotimality in the decoder’s output. Another advantage of this technique is that one can further collect various paraphrases by exploiting multiple languages.  

\section{Experiments}
\subsection{Data}
We evaluate our English model on TurkCorpus \cite{xu-etal-2016-optimizing} and PWKP \cite{zhang-lapata-2017-sentence}. TurkCorpus is a standard dataset for evaluating sentence simplification works, originally extracted from WikiLarge corpus compiled from \cite{zhang-lapata-2017-sentence}. It contains 2000 sentences for validation and 359 for testing, each has 8 simplification references collected through crowd-sourcing. PWKP is the test set of WikiSmall - another dataset constructed from main-simple Wikipedia articles. PWKP provides 100 test sentences with 1-to-1 aligned reference. Newsela \cite{xu-etal-2015-problems} is another commonly used dataset in text simplification works, which is unfortunately unavailable due to restricted access rights. Without training, we directly run our model and evaluate the outputs on TurkCorpus and PWKP test sets, and to prove the robustness of our system, we further use CNN/Daily Mail, a different dataset \cite{see-etal-2017-get} for training the fluency model, leaving both evaluation corpora untouched. For the Vietnamese model, we train the fluency model on the public Vietnamese news corpus CP\_Vietnamese-UNC of 41947 sentences, then generate simplifications of 200 sentences extracted from Vietnamese law corpus CP\_Vietnamese-VLC in an unsupervised manner. Both datasets are open sourced by Underthesea NLP \footnote{\url{github.com/undertheseanlp/resources}}. 

\subsection{System Details}
We utilize SpaCy\footnote{\url{spacy.io}} and Berkeley Neural Parser \cite{kitaev-klein-2018-constituency} for constituent and dependency parsing. Since SpaCy does not support Vietnamese, we parse Vietnamese texts through VnCoreNLP\footnote{\url{github.com/vncorenlp/VnCoreNLP}} \cite{vu-etal-2018-vncorenlp}. VnCoreNLP is built upon Vietnamese Treebank \cite{nguyen-etal-2009-building}, which contains 10,200 constituent trees formatted similarly to Penn Treebank \cite{marcus-etal-1993-building}. We obtain sentence embeddings from SBERT model \texttt{paraphrase-mpnet-base-v2}  \cite{reimers-2020-multilingual-sentence-bert} for English, and the multilingual version \texttt{distiluse-base-multilingual-cased -v2} for Vietnamese \cite{reimers-gurevych-2020-making}. Our fluency model is a 4-gram language model with Kneser-Ney smoothing implemented via NLTK\footnote{\url{nltk.org}} package. Before that, we randomly sample 1 million sentences from CNN/Daily Mail set and parse them into sequences of constituents. For example, the sentence \textit{Their eyes are small} is transformed into \texttt{{PRON NOUN AUX ADJ}}.  

The back-translation procedure employs free Google Translate service\footnote{\url{translate.google.com}} - a robust neural translation system that support two-way translation {\bf English-German} and {\bf German-English}. Though the system can be run with any available target languages, German is selected for our English model because it is a well-resourced language. For the Vietnamese model, we simply use English as the target language. While our search algorithm can automatically guarantee language idiomaticity, grammatical accuracy is an issue since the model tends to prefer content words to function words to maximize semantic similarity. Thus, we strictly force Fluency to be twice as important, i.e., $\alpha=2$, while set equal weights to Similarity and Depth constraint. Our base model is evaluated at $\lambda=0.5$ and $\tau=0.95$,  reported as \texttt{USDP-Base}. In order to validate our effectiveness more comparably, we produce additional variants approximating $\tau$ to same level of competing unsupervised methods, respectively at $\tau=0.90$ for TurkCorpus (\texttt{USDP-Match$^{a}$}) and $\tau=0.75$ for PWKP (\texttt{USDP-Match$^{b}$}). Across all experiments, \textbf{beam size is fixed at 5}, and in order to understand the effect of back-translation, we also evaluate the quality of simplifications before and after phase $2$.  
\subsection{Competing Models}
We benchmark our system against existing supervised and unsupervised English sentence simplification models. Supervised systems include \texttt{PBMT-R} \cite{wubben-etal-2012-sentence}, \texttt{SBMT-SARI} \cite{xu-etal-2016-optimizing}, \texttt{Dress} / \texttt{Dress-Ls} \cite{zhang-lapata-2017-sentence} and recent state-of-the-art \texttt{ACCESS} \cite{martin-etal-2020-controllable}. We also consider semi-supervised \texttt{BTTS / BTRLTS / BTTS100} \cite{zhao2020semi} and unsupervised counterparts \texttt{UNTS / UNTS10K} \cite{surya-etal-2019-unsupervised} and \texttt{RM+EX / RM+EX+LS / RM+EX+RO / RM+EX+LS+RO} \cite{kumar-etal-2020-iterative}.
\section{Results}
\subsection{Automatic Evaluation}
We use the EASSE package \cite{alva-manchego-etal-2019-easse, martin-etal-2018-reference} to compute standardized simplification metrics and perform evaluation on publicly accessible outputs of competing systems. These include Compression ratio (\textbf{CR}), Exact copies (\textbf{CP}), Split ratio (\textbf{\%SP}), Additions proportion (\textbf{\%A}) and Deletions proportion (\textbf{\%D}), all of which are evaluated against the source sentences. Details on these automatic metrics can be found in Appendix \ref{appendixA}. We exclude BLUE and FKGL since BLEU is previously reported to be a poor estimate of simplicity and FKGL only applies to text of at least 200 words \cite{alva-manchego-etal-2020-asset,wubben-etal-2012-sentence,xu-etal-2016-optimizing}. We add to the current simplification evaluation suite the measures of semantic similarity and fluency. Similarity score is again based on cosine similarity between sentence embedding vectors (\textbf{SIM}), and we adopt a “third-party” language model for scoring fluency (\textbf{FL}). This is to assure fair comparison among systems since ours has a different fluency scoring scheme. We use \textbf{pseudo-log-likelihood scores} (PPLs) proposed in \cite{salazar-etal-2020-masked}, which is shown to promote linguistic fluency rather than pure likelihood in conventional log probabilities. We also evaluate the reference sentences on these quality metrics, and benchmark the system outputs against them through average \textbf{SARI} and component \textbf{Add}, \textbf{Keep}, \textbf{Del} scores \cite{xu-etal-2016-optimizing}. This however is done only on TurkCorpus set since PWKP only provides 1 reference. Table \ref{tab2}, \ref{tab3} and \ref{tab4} present results of automatic evaluation respectively on TurkCorpus, PWKP and CP\_Vietnamese-VLC, both before and after back-translation (\textbf{BT}). 

\subsubsection{TurkCorpus} We establish the unsupervised state-of-the-art SARI on TurkCorpus with \textbf{+1.65} point improvement over the closest baseline and only behind two supervised methods: \texttt{ACCESS} and \texttt{SBMT-SARI}. In addition to the competitive performance on Compression ratio and Split ratio, we outperform the current semi-supervised and unsupervised across all other quality metrics. Our simplifications have the highest fluency at \textbf{-2.47} and similarity score at \textbf{0.95} while achieving remarkably high percentages of additions at \textbf{16\%} and deletions of \textbf{21-25\%} at the same level of some supervised methods. Our raw outputs from phase 1 alone gains fairly high proportions of deletions as lowering $\tau$. For other methods, this number generally takes both deletions and substitutions into account, but in this phase, it reflects our model's effectiveness in performing deletions since substitutions are not implemented until phase 2. Figures \ref{fig4} and \ref{fig5} in Appendix \ref{appendixC} visualize how well our base model balances simplicity with adequacy and fluency compared to other methods. Samples of system outputs are additionally provided in Appendix \ref{appendixD}.   

\begin{table*}[hbt!]
\centering
\resizebox{0.8\textwidth}{!}{%
\begin{tabular}{l|l l l l l|l l|l l l l}
\hline
\textbf{TurkCorpus} & \textbf{CR$\downarrow$} & \textbf{CP$\downarrow$} & \textbf{\%SP$\uparrow$} & \textbf{\%A$\uparrow$} & \textbf{\%D$\uparrow$} & \textbf{FL$\uparrow$} & \textbf{SIM$\uparrow$} & \textbf{SARI$\uparrow$} & \textbf{Add$\uparrow$} & \textbf{Keep$\uparrow$} & \textbf{Del$\uparrow$} \\
\hline
\texttt{Reference}  & 0.95 & 1.07  & 0.16     & 0.14 & 0.18  & -2.63  & \textbf{0.95} & - & - & - & - \\
\hline
\multicolumn{12}{l}{\textbf{Supervised}} \\ \hline
\texttt{Dress}       & \textbf{0.75}          & 0.22          & 0.99      & 0.04          & \textbf{0.27} & -2.66          & 0.91          & 36.84          & 2.5           & 65.65          & 42.36          \\
\texttt{Dress-Ls}    & \textbf{0.77}          & 0.26          & 0.99      & 0.04          & \textbf{0.26} & -2.63          & 0.92          & 36.97          & 2.35          & 67.23          & 41.33          \\
\texttt{PBMT-R}      & 0.95          & 0.11          & 1.03      & 0.10          & 0.11          & -2.59          & \textbf{0.96} & \textbf{38.04} & 5.04          & \textbf{73.77} & 35.32          \\
\texttt{ACCESS}      & 0.94          & \textbf{0.04} & \textbf{1.20}       & \textbf{0.16} & 0.16          & -2.52          & \textbf{0.95} & \textbf{41.38} & \textbf{6.58}          & 72.79          & 44.78          \\
\texttt{SBMT-SARI}   & 0.94          & 0.10          & 1.02      & \textbf{0.16} & 0.13          & -2.65          & \textbf{0.96} & \textbf{39.56} & 5.46          & 72.44          & 40.76          \\ \hline
\multicolumn{12}{l}{\textbf{Semi-Supervised}} \\\hline

\texttt{BTTS100}     & 0.92          & 0.45          & 1.02      & 0.03          & 0.10          & \textbf{-2.46} & \textbf{0.97} & 34.48          & 1.51          & \textbf{74.44} & 27.48          \\
\texttt{BTTS}        & 0.92          & 0.20          & 1.17      & 0.08          & 0.14          & -2.66          & \textbf{0.96} & 36.38          & 1.9           & 71.03          & 36.22          \\
\texttt{BTRLTS}      & 0.92          & 0.19          & 1.16      & 0.08          & 0.15          & -2.70          & \textbf{0.96} & 36.49          & 2.14          & 70.31          & 37.03          \\
\hline
\multicolumn{12}{l}{\textbf{Unsupervised}} \\ \hline
\texttt{UNTS}        & 0.85          & 0.21          &  1.00          & 0.06          & 0.17          & -2.70          & 0.89          & 36.29          & 0.83          & 69.44          & 38.61          \\
\texttt{UNTS\_10K}   & 0.88          & 0.19          & 1.01      & 0.07          & 0.14          & -3.10           & 0.92          & 37.15          & 1.12          & 71.34          & 38.99          \\
\texttt{RM+EX}       & 0.83          & 0.44          &  1.00          & 0.01          & 0.15          & -2.58          & 0.94          & 35.88          & 0.84          & 73.14          & 33.65          \\
\texttt{RM+EX+LS}    & 0.82          & 0.16          &  1.00          & 0.06          & \textbf{0.21}          & -2.91          & 0.90         & 37.48          & 1.59          & 68.20           & 42.65          \\
\texttt{RM+EX+RO}    & 0.86          & 0.36          & 1.01      & 0.02          & 0.14          & -2.61          & 0.94          & 36.07          & 0.99          & 72.36          & 34.86          \\
\texttt{RM+EX+LS+RO} & 0.85          & 0.13          & 1.01      & 0.08   &       \textbf{0.20}          & -2.92          & 0.90         & 37.27          & 1.68          & 67.00             & 43.12          \\ \hline
\multicolumn{12}{l}{\textbf{Our system}} \\
\hline
\multicolumn{12}{l}{\texttt{USDP-Base}} \\

\textbf{With BT}     & 0.92          & \textbf{0.04} & 1.01      & \textbf{0.16} & \textbf{0.21} & \textbf{-2.47} & \textbf{0.95} & \textbf{39.13} & \textbf{6.77} & 64.44          & \textbf{46.19} \\
\textbf{Without BT}  & 0.95          & 0.15          &  1.00          & 0.07          & 0.09          & -2.88         & \textbf{0.98} & 34.13          & 0.87          & 71.34          & 30.18          \\ [1ex]

\multicolumn{12}{l}{\texttt{USDP-Match$^{a}$}} \\

\textbf{With BT}     & 0.88          & \textbf{0.04} & 1.01      & \textbf{0.15} & \textbf{0.25} & -2.55          & 0.94          & \textbf{38.33} & \textbf{6.28} & 62.13          & \textbf{46.58} \\
\textbf{Without BT}  & 0.89          & 0.13          &  1.00          & 0.07          & 0.15          & -3.05         & \textbf{0.96} & 34.44          & 0.94          & 68.57          & 33.82 \\        

\hline
\end{tabular}
}%
\caption{Results on TurkCorpus. $\uparrow$ Higher is better. $\downarrow$ Lower is better.} \label{tab2}
\end{table*}

\subsubsection{PWKP} As \citet{kumar-etal-2020-iterative} do not experiment on PWKP, we run their codes to evaluate \texttt{RM+EX+LS+RO} model on PWKP set for comparison. Overall, our model and \texttt{RM+EX+LS+RO} produce more diverse simplifications than the supervised systems, measured by significantly higher proportion of additions and deletions. Interestingly, the quality of unsupervised outputs is also closer to that of reference outputs, in which \texttt{RM+EX+LS+RO} achieves consistently better performance. This is probably because the model is accompanied by a pre-trained Word2Vec on WikiLarge data, which has a relatively similar distribution to PWKP as both are Wikipedia-based. Meanwhile, none of our variants see any similar examples of any kind beforehand. Given such a high level of modification, we again have the highest similarity score at \textbf{0.96}, and when we try to match the similarity level as in \texttt{USDP-Match$^{b}$}, we achieve more compression at \textbf{54\%} and deletions at \textbf{56\%} while preserving slightly higher semantics than \texttt{RM+EX+LS+RO}. The fluency scores of outputs from all automated systems remain far behind human simplifications.  

\begin{table}[h!]
\centering
\resizebox{0.50\textwidth}{!}{%
\begin{tabular}{l|c c c c c|c c}
\hline
\textbf{PWKP} & \textbf{CR$\downarrow$} & \textbf{CP$\downarrow$} & \textbf{\%SP$\uparrow$} & \textbf{\%A$\uparrow$} & \textbf{\%D$\uparrow$} & \textbf{FL$\uparrow$} & \textbf{SIM$\uparrow$} \\
\hline
\texttt{Reference}    & 0.81          & 0.03          & \textbf{1.31} & \textbf{0.17} & 0.32          & \textbf{-1.39} & 0.91 \\
\hline
\multicolumn{8}{l}{\textbf{Supervised}} \\ \hline
\texttt{Dress}        & 0.62          & 0.11          & 1.01          & 0.02          & 0.39          & -2.18          & 0.87 \\
\texttt{Dress-Ls}     & 0.63          & 0.13          & 1.01          & 0.01          & 0.37          & -2.10          & 0.88 \\
\texttt{PBMT-R}       & 0.96          & 0.14          & 1.01          & 0.06          & 0.07          & -2.05          & \textbf{0.97} \\\hline
\multicolumn{8}{l}{\textbf{Unsupervised}} \\ \hline
 \texttt{RM+EX+LS+RO}  & 0.61         & \textbf{0.01} & \textbf{1.21} & \textbf{0.17} & \textbf{0.52} & -2.68          & 0.81 \\\hline
\multicolumn{8}{l}{\textbf{Our system}} \\ \hline
\multicolumn{8}{l}{\texttt{USDP-Base}}\\
 \textbf{With BT}   & 0.87          & 0.03  & 1.00  & \textbf{0.16} & 0.28          & -2.05       & \textbf{0.96} \\
\textbf{Without BT}  & 0.88         & 0.08  & 1.00  & 0.06          & 0.15          & -2.64 & \textbf{0.95} \\ [1ex]
\multicolumn{8}{l}{\texttt{USDP-Match$^{b}$}} \\
\textbf{With BT}      & \textbf{0.54} & \textbf{0.00} & 1.00          & 0.11          & \textbf{0.56} & -2.38          & 0.84 \\
\textbf{Without BT}   & \textbf{0.53} & 0.03          & 1.00          & 0.03          & 0.49          & -3.36          & 0.85 \\
\hline
\end{tabular}
}
\caption{Results on PWKP. $\uparrow$ Higher is better. $\downarrow$ Lower is better.} \label{tab3}
\end{table}
\subsubsection{CP\_Vietnamese-VLC}
As a proof-of-concept for our approach in another language, we only conduct evaluation on \texttt{USDP-Base}. We do not report PPLs since that model has only been shown to work on English data. We instead report \textbf{LevSIM}, normalized character-level Levenshtein similarity \cite{Levenshtein_SPD66}, which demonstrates that the output structures do not deviate significantly from the original. Results in Table \ref{tab4} are consistent with what we have achieved on English corpora, proving the potential to apply the framework to other languages.    

\subsection{Human Evaluation}
Human judgement is critical to assess text generation. We randomly select 50 sentences from TurkCorpus test set, and have 5 volunteers (2 native and 3 non-native speakers with adequate English proficiency) examine the simplified outputs from \texttt{ACCESS} (supervised state-of-the-art), \texttt{RM+EX+LS} (closest and best performing unsupervised variant) and our method \texttt{USDP-Base}. In a similar setup to the previous studies \cite{kumar-etal-2020-iterative,NAACL-2021-Maddela,zhao2020semi}, the volunteers are asked to provide ratings for each simplification version on 3 dimensions: \textbf{Simplicity}, \textbf{Adequacy} and \textbf{Fluency}. We also have 50 Vietnamese simplifications from CP\_Vietnamese-VLC outputs assessed by 4 native Vietnamese speakers on the same quality dimensions.  We simply report the average ratings in Table \ref{tab5}, substantiating that we surpass both \texttt{ACCESS} and \texttt{RM+EX+LS} on all dimensions, and our simplified sentences in Vietnamese are perceived to have adequate quality.

\begin{table*}[h!]
\centering
\resizebox{0.7\textwidth}{!}{%
\begin{tabular}{l|l l l l l|l l}
\hline
\textbf{CP\_Vietnamese-VLC} & \textbf{CR$\downarrow$} & \textbf{CP$\downarrow$} & \textbf{\%SP$\uparrow$} & \textbf{\%A$\uparrow$} & \textbf{\%D$\uparrow$} & \textbf{LevSIM$\uparrow$} & \textbf{SIM$\uparrow$} \\
\hline
\textbf{With BT} & 0.89  & 0.00  & 1.06      & 0.11 & 0.20 & 0.86  & 0.91 \\
\textbf{Without BT} & 0.91  & 0.00  & 0.99  & 0.06 & 0.12 & 0.91  & 0.94 \\
\hline
\end{tabular}
}
\caption{Results of \texttt{USDP-Base} on CP\_Vietnamese-VLC}\label{tab4}
\end{table*}

\begin{table}[h!]
\centering
\resizebox{0.35\textwidth}{!}{%
\begin{tabular}{p{2cm}|c|c|c}
\hline
\textbf{Model}      & \textbf{Fluent}  & \textbf{Adequate}     & \textbf{Simple} \\ [0.5ex]
\hline
\multicolumn{4}{l}{\textbf{English}} \\ \hline
\texttt{USDP-Base}  & \textbf{4.32}     & \textbf{3.93}         & \textbf{3.22} \\
\texttt{ACCESS}	    & 4.16	            & 3.46	                & 3.18 \\
\texttt{RM+EX+LS}	& 3.59	            & 3.12	                & 2.86 \\
\hline
\multicolumn{4}{l}{\textbf{Vietnamese}} \\ \hline
\texttt{USDP-Base}  & 3.33     & 3.48         & 3.04 \\
\hline
\end{tabular}
}
\caption{Human Evaluation Results on TurkCorpus (English) and CP\_Vietnamese-VLC (Vietnamese) datasets.}
\label{tab5}
\end{table}

\subsection{Controllability}
Table \ref{tab6} displays output results on different values of $\tau$ and $\lambda$. Adjusting similarity threshold $\tau$ has more impact on the output quality than length ratio $\lambda$. This is simply because our algorithm must satisfy a pre-defined $\tau$ before termination, regardless of length constraints. This shows that lowering similarity threshold encourages the model to produce more deletions and compression, which however does not occur at the cost of semantics preservation. Even when $\tau$ is set to $0.70$, the output sentences still have very high similarity scores. Little content is lost since we not only reduce length but also seek to maximize semantics preservation by  extracting important tokens only, most of which turn out to be content words. This behavior is discussed in detail in the next section.

\begin{table}
\centering
\resizebox{0.35\textwidth}{!}{%
\begin{tabular}{ p{1cm}|p{1cm}|p{1cm}|p{1cm}|p{1cm}  }
\hline
\textbf{Value} & \textbf{CR$\downarrow$} & \textbf{\%D$\uparrow$} & \textbf{SARI$\uparrow$} & \textbf{SIM$\uparrow$} \\ \hline
\multicolumn{5}{c}{Effect of $\lambda$ at $\tau=0.95$} \\ \hline 
$0.25$         & 0.95       & 0.08          & 33.60        & 0.98        \\
$0.50$         & 0.95       & 0.08          & 33.60        & 0.98        \\
$0.75$         & 0.96       & 0.07          & 33.47        & 0.98        \\
$1.00$          & 0.99       & 0.04          & 32.86        & 0.98        \\[0.5ex] \hline
\multicolumn{5}{c}{Effect of $\tau$ at $\lambda=0.5$} \\ \hline
$0.70$         & 0.81       & 0.22         & 33.63       & 0.92        \\
$0.80$         & 0.84       & 0.19          & 33.77        & 0.94        \\
$0.90$         & 0.91       & 0.12          & 33.80        & 0.97        \\
$0.95$         & 0.95       & 0.08          & 33.60        & 0.98 \\ \hline
\end{tabular}
}
\caption{Effects of threshold values on simplification quality of $100$ sentences from TurkCorpus. Both are evaluated on \texttt{USDP-Base}.}\label{tab6}
\end{table}

\section{Discussion}
In the first phase, we ensure each added token brings about significant improvement in semantics. A syntactic investigation was conducted to understand this behavior better, illustrated in Figure \ref{fig6} in Appendix \ref{appendixC}. Figure \ref{fig6}$a$ first examines the correlation between the tokens' part-of-speech and changes in similarity on average. After randomly sampling $1$ million English sentences from all the data, for each sentence, we consecutively remove every token and track how much the SBERT embedding similarity score is reduced, then add the token back in before evaluating a new one. \textbf{Content words} including \texttt{NOUN}, \texttt{PRON}, \texttt{VERB}, \texttt{ADJ} and \texttt{ADV}, which are deemed semantically more important, each causes more than $6\%$ reduction in semantic value when omitted. Meanwhile, \textbf{function words} such as \texttt{CONJ} or \texttt{DET} only reduces similarity scores by less than $4\%$ each. This implies that a child token serving as an adposition, for instance, is less likely to be selected in the sampling set compared to when it is a noun or adjective, especially in the family with more members than the chosen beam size. Hence, using the family sampling algorithm, the decoder is likely to eliminate the entire branch corresponding to prepositional or adverbial phrases in the dependency tree, as reported in Appendix \ref{appendixD}. We also examine the effect of tree depth and sentence length (word count) on similarity changes, which is discussed in \cite{schumann-etal-2020-discrete} as a problem of position bias. We find that this is not a major issue to our work. Although a large portion of important content is allocated towards the top of the parsing tree as well as the beginning of the sentence (Figure \ref{fig6}$b$), the distributions of content words and function words are almost uniform across the tree (Figure \ref{fig6}$c$). We therefore rule out position bias and instead attribute this to human nature of writing. 

With respect to the second phase, the role of back-translation is to introduce meaningful diversity, and we observe that back-translation does more paraphrasing than simple lexical substitution. Therefore, sometimes the output sentences must be longer to be re-written in a simpler way, resulting in slightly less compression (i.e., relatively high Compression ratio). 

\section{Conclusion and Future Work}
We implement the novel \textbf{family sampling} strategy on top of the regular beam-search-based decoding for sentence simplification. Our work directly tackles the data scarcity issue by proposing an unsupervised framework that generates hybrid outputs in a light-weight architecture. 

Our mechanism can be adapted to paragraph-level simplification by measuring semantic changes to the paragraph given the removal of any sentence. In this case, the importance of a sentence should also factor in the co-referent relation with other sentences as proposed in \cite{liu2021unsupervised}. Since we already have phrasal chunks, we can additionally investigate where reordering or deleting any of them would result in a simpler tree structure without significant semantic reduction. Furthermore, our proof-of-concept of Vietnamese simplification demonstrates it has scope for improvement, and that the framework can also be applied to other languages with similar dependency structures.

\bibliography{anthology,custom}
\bibliographystyle{acl_natbib}

\appendix
\section{Details of Automatic Metrics} \label{appendixA}
This section explains the computation of automatic evaluation metrics provided by EASSE \cite{alva-manchego-etal-2019-easse}:
\begin{itemize}
  \item \textbf{Compression ratio (CR):} average ratio of number of characters in the output to that in the original.
  \item \textbf{Exact Copies (CP):} proportion of outputs exactly same to the original word-wise. 
  \item \textbf{Split ratio (\%SP):} average ratio of number of sentences in the output to that in the original.
  \item \textbf{Additions proportion (\%A):} average proportion of words in the output but not in the original. 
  \item \textbf{Deletions proportion (\%D):} average proportion of words removed from the original.
  \item \textbf{SARI:} for each operation $ope \in \{add,del,keep\}$, $n-$gram and its order between the output and all references, we calculate precision $p_{ope}(n)$, recall $r_{ope}(n)$ and F1 score $f_{ope}(n)$ by 
  $$f_{ope}(n) = \frac{2 \times p_{ope}(n) \times r_{ope}(n)}{p_{ope}(n) + r_{ope}(n)}$$
  Averaging over the $n-$gram orders ($k$), the overall operation F1 score is 
  $$F_{ope} = \frac{1}{k}\sum f_{ope}(n)$$
  $n=4$ is a popular choice.
\end{itemize}

\section{Details of Human Evaluation} \label{appendixB}
A human evaluation sheet consists of instructions followed by 50 blocks of texts. Each block contains 1 original sentence and 3 simpler variants corresponding to the simplified output from each system. The order of the variants within each block is also randomized, and we do not annotate which output belongs to which system (i.e., blind evaluation). The participants are required to use a five-point Likert scale to rate their degree of agreement to the following statements
\begin{itemize}
  \item \textbf{Simplicity:} The output is simpler than the original sentence.
  \item \textbf{Adequacy: } The meaning expressed in the original sentence is preserved in the output.
  \item \textbf{Fluency:} The output sentence is grammatical and well formed.

\end{itemize}
\begin{figure*}[h!]
\centering
\includegraphics[scale=0.45]{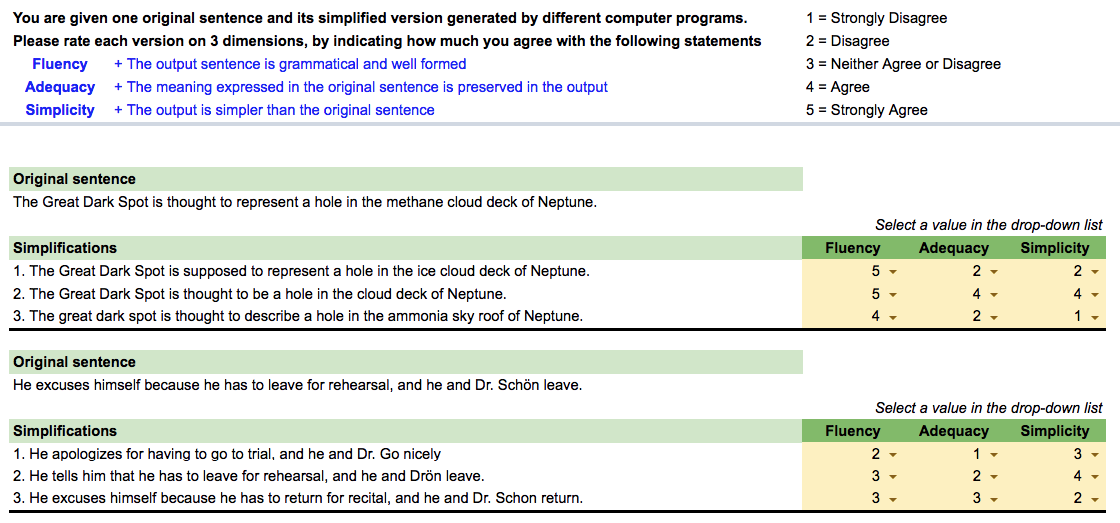}
\caption{English Human Evaluation Interface}
\label{fig2}
\end{figure*}

\begin{figure*}[h!]
\centering
\includegraphics[scale=0.45]{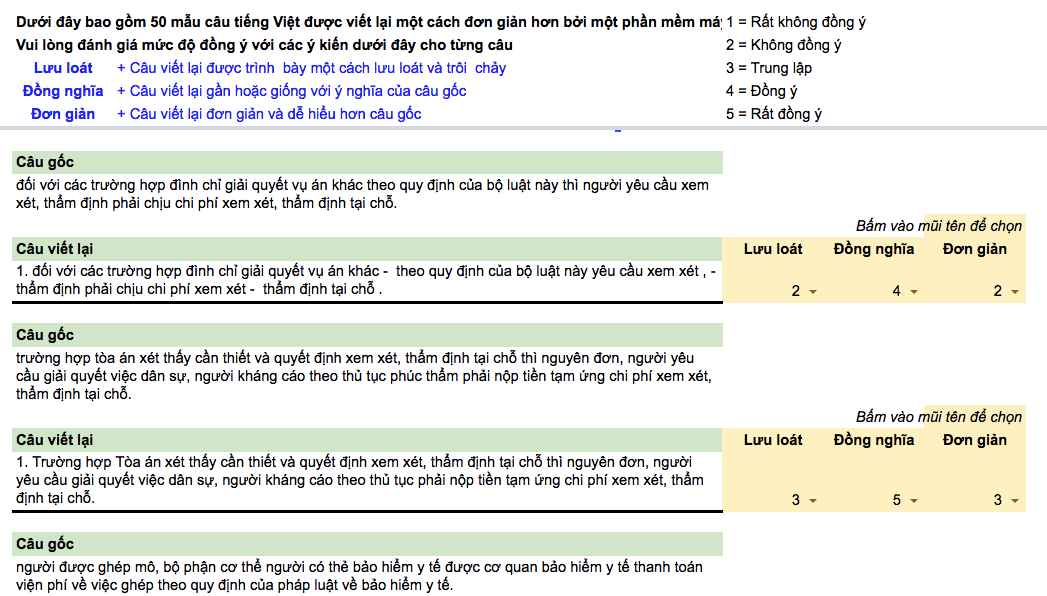}
\caption{Vietnamese Human Evaluation Interface}
\label{fig3}
\end{figure*}

\clearpage
\begin{figure*}
\section{System Effectiveness} \label{appendixC}
\centering
\includegraphics[width=1.0\textwidth]{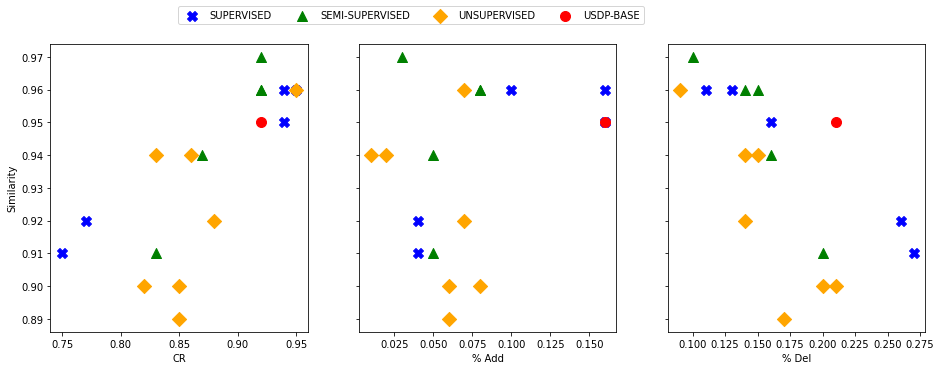}
\caption{Visualization of systems' capacity to balance \textbf{Adequacy} with \textbf{Simplicity} on TurkCorpus}
\label{fig4}
\end{figure*}

\begin{figure*}
\centering
\includegraphics[width=1.0\textwidth]{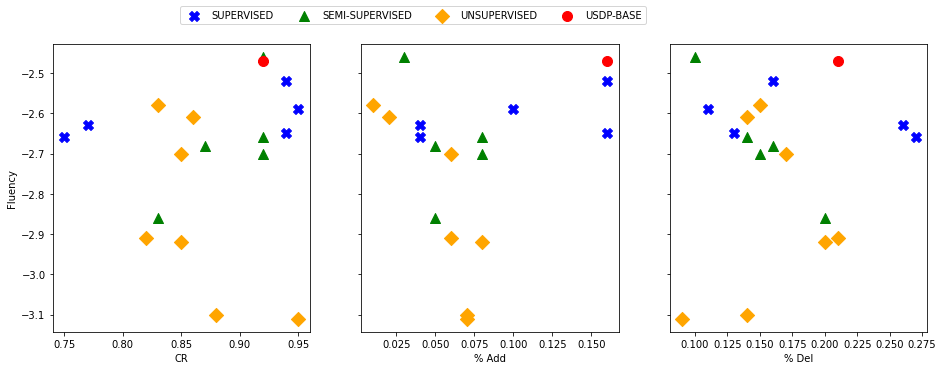}
\caption{Visualization of systems' capacity to balance \textbf{Fluency} with \textbf{Simplicity} on TurkCorpus}
\label{fig5}
\end{figure*}

\begin{figure*}
\centering
\includegraphics[width=1.0\textwidth]{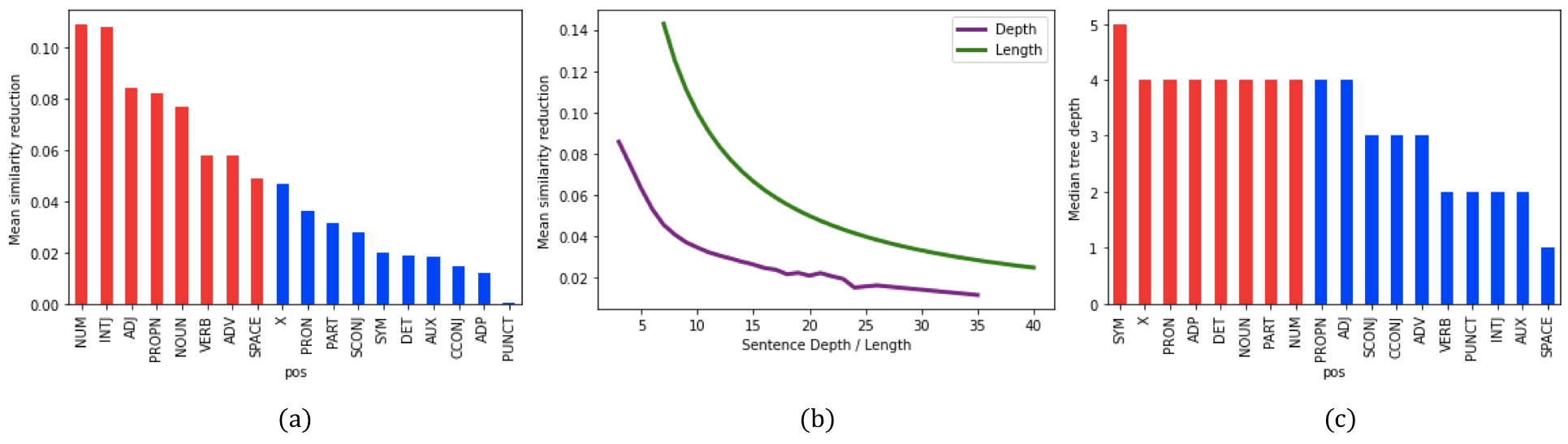}
\caption{Figure \textbf{(a)} shows the effect of grammatical functionality (part-of-speech) of words on reduction in similarity scores. \textbf{(b)} illustrates the correlation between location of words and reduction in similarity scores i.e., the allocation of important words in the sentence. \textbf{(c)} displays the uniform distribution of part-of-speeches across the sentence. }
\label{fig6}
\end{figure*}

\begin{table*}[h]
\section{Qualitative Evaluation} \label{appendixD}
\setlength\tabcolsep{3 pt}
\begin{tabular}{l|p{0.85\linewidth}}
\hline
\multicolumn{2}{l}{\textbf{Example of deleting prepositional/adjective phrases}}  \\
\hline
Original sentence & \textit{Jeddah is the \textbf{\textcolor{blue}{principal}} gateway to Mecca, Islam's holiest city, which able-bodied Muslims \textbf{\textcolor{blue}{are required to visit at least once}} in their lifetime.}  \\
\multicolumn{2}{l}{System outputs} \\ 
\texttt{RM+EX+LS}   & \textit{Jeddah is the principal gateway to Mecca, Islam's Holiest city, which Able-Bodied Muslims are required to visit at least once in their lifetime.} \\ [0.5ex]
\texttt{USDP-Base}         &    \\ [0.5ex]
\textbf{With BT} & \textit{Jeddah is the \textbf{\textcolor{blue}{main}} gateway to Mecca, the holiest city in Islam that Muslim people \textbf{\textcolor{blue}{had to visit}} during their lifetime.} \\ [0.5ex]
\textbf{Without BT} & \textit{Jeddah is the principal gateway to Mecca , Islam 's holiest city , which able - bodied Muslims required -  visit in their lifetime.} \\ [0.5ex]
\hline
\multicolumn{2}{l}{\textbf{Example of summarization}} \\ 
\hline
Original sentence & \textit{At four-and-a-half years old he was left to fend for himself on the streets of northern Italy for the next four years, \textbf{\textcolor{blue}{living in various orphanages and roving through towns with groups of other homeless children.}}} \\ [0.5ex]
\multicolumn{2}{l}{System outputs} \\ [0.5ex] 
\texttt{RM+ES+LS}          & \textit{At Four-And-A-Half years old he was left to fend for himself on the walls of northern Italy for the next four years.} \\ [0.5ex]
\texttt{USDP-Base}         &  \\ [0.5ex]
\textbf{With BT}           & \textit{At the age of four and a half, he had to support himself on the streets of northern Italy for the next four years, \textbf{\textcolor{blue}{wandering through the cities living in various orphanages.}}}   \\ [0.5ex]
\textbf{Without BT}        & \textit{At four - and - a - half years old he was left -  to fend for himself on the streets of northern Italy for the next four years , living in various orphanages -  roving through towns.}.  \\ [0.5ex]
\hline
\multicolumn{2}{l}{\textbf{Example of chunking}} \\
\hline
Original sentence & \textit{In late 2004, Suleman made headlines by cutting Howard Stern's radio show from four Citadel stations, citing Stern's frequent discussions regarding his upcoming move to Sirius Satellite Radio.}  \\ [0.5ex]
\multicolumn{2}{l}{System outputs} \\ [0.5ex]                   \texttt{RM+ES+LS}          & \textit{In late 2004, Suleman made headlines by cutting Howard Stern'S radio show from four Citadel trains, reporting Stern'S serious questions.} \\ [0.5ex]
\texttt{USDP-Base}         & \\ [0.5ex]
\textbf{With BT}           & \textit{In late 2004, Suleman made headlines - \textbf{\textcolor{blue}{by cutting Howard Stern's radio show from four Citadel stations} - \textbf{\textcolor{blue}{citing Stern's discussions}} - \textbf{\textcolor{blue}{regarding the upcoming move}}.}} \\ [0.5ex]
\textbf{Without BT}        & \textit{In late 2004, Suleman made headlines - by cutting Howard Stern's radio show from four Citadel stations - citing Stern's discussions - regarding upcoming move.} \\ [0.5ex]
\hline
\multicolumn{2}{l}{\textbf{Example of simplistic paraphrasing}} \\
\hline
Original sentence & \textit{Fearing that Drek will destroy the galaxy, Clank asks Ratchet to \textbf{\textcolor{blue}{help him find the famous superhero Captain Qwark, in an effort to stop Drek.}}} \\ [0.5ex]
\multicolumn{2}{l}{System outputs} \\ [0.5ex]  
\texttt{RM+ES+LS}  & \textit{Fearing that Drek will bring the universe, Clank asks Ratchet to help him find the famous Superhero captain Qwark, in an attempt to get Drek.} \\ [0.5ex]
\texttt{USDP-Base}         & \\ [0.5ex]
\textbf{With BT}           & \textit{Fearing Drek might destroy the galaxy, Clank asks Ratchet to \textbf{\textcolor{blue}{find the superhero in order to stop Drek}}.}  \\ [0.5ex]
\textbf{Without BT}      & \textit{Fearing that Drek will destroy the galaxy , Clank asks Ratchet -  help find the superhero -  in effort stop Drek.} \\ [0.5ex] \hline
\end{tabular}
\caption{Qualitative results on TurkCorpus.}\label{tab7}
\end{table*}

\end{document}